\documentclass[10pt,twocolumn,letterpaper]{article}

\usepackage{cvpr}
\usepackage{times}
\usepackage{epsfig}
\usepackage{graphicx}
\usepackage{amsmath}
\usepackage{amssymb}
\usepackage{multirow}
\usepackage{mathrsfs}
\usepackage{subfig}

\usepackage[pagebackref=true,breaklinks=true,letterpaper=true,colorlinks,bookmarks=false]{hyperref}

\cvprfinalcopy 

\def\cvprPaperID{3351} 

\ifcvprfinal\fi
\begin{document}

\title{VRSTC: Occlusion-Free Video Person Re-Identification}

\author{Ruibing Hou$^{1,2}$, Bingpeng Ma$^{2}$, Hong Chang$^{1,2}$, Xinqian Gu$^{1,2}$, Shiguang Shan$^{1,2,3}$, Xilin Chen$^{1,2}$\\
$^1$Key Laboratory of Intelligent Information Processing of Chinese Academy of Sciences (CAS),\\Institute of Computing Technology, CAS, Beijing, 100190, China\\
$^2$School of Computer and Control Engineering, \\University of Chinese Academy of Sciences, Beijing, 100049, China\\
$^3$CAS Center for Excellence in Brain Science and Intelligence Technology, Shanghai, 200031, China\\
{\tt\small \{ruibing.hou, xinqian.gu\}@vipl.ict.ac.cn, bpma@ucas.ac.cn, \{changhong, sgshan,xlchen\}@ict.ac.cn}
}

\maketitle

\begin{abstract}
Video person re-identification (re-ID) plays an important role in surveillance video analysis. However, the performance of video re-ID degenerates severely under partial occlusion. In this paper, we propose a novel network, called Spatio-Temporal Completion network (STCnet), to explicitly handle partial occlusion problem. Different from most previous works that discard the occluded frames, STCnet can recover the appearance of the occluded parts. For one thing, the spatial structure of a pedestrian frame can be used to predict the occluded body parts from the unoccluded body parts of this frame. For another, the temporal patterns of pedestrian sequence provide important clues to generate the contents of occluded parts. With the spatio-temporal information, STCnet can recover the appearance for the occluded parts, which can be leveraged with those unoccluded parts for more accurate video re-ID. By combining a re-ID network with STCnet, a video re-ID framework robust to partial occlusion (VRSTC) is proposed. Experiments on three challenging video re-ID databases demonstrate that the proposed approach outperforms the state-of-the-art approaches.
\end{abstract}

\section{Introduction}
Video person re-identification (re-ID) aims at matching the same person across multiple non-overlapping cameras, which has gained increasing attention in recent years. However, it remains a very challenging problem due to large variations of appearance caused by camera viewpoints, background clutter, and especially partial occlusion. The performance of video re-ID usually degenerates severely under partial occlusion. This problem is difficult to tackle as any part of the person may be occluded by other pedestrians and environmental objects (\eg bicycles and indicators). 

Typical video re-ID methods \cite{RCN,re-conv,temporal-enhanced} do not take into account the effect of partial occlusion. They represent each frame of a video as a feature vector and compute an aggregate representation across time with average or maximum pooling. In the presence of partial occlusion, the video feature is usually corrupted due to the equal treatment of all frames, leading to severe performance degeneration.

Recently, the attention mechanism has been introduced to video re-ID in order to deal with partial occlusion \cite{QAN,See,jointly,diversity,snippet}. They select discriminative frames from video sequences and generate informative video representation. Although these approaches have a certain tolerance to partial occlusion, it is not ideal to discard the occluded frames. On one hand, the remaining visible portions of the discarded frames may provide strong cues for re-ID. So these methods lost too much appearance information in video features, making them difficult to identify the person. On the other hand, the discarded frames interrupt the temporal information of video. The works \cite{RCN,re-conv,temporal-enhanced} have verified that the temporal information of video can help to identify the person. For instance, if different persons have similar appearance, we can disambiguate them from their gaits. Therefore, these methods may still fail when the partial occlusion occurs.

In this work, we propose Spatial-Temporal Completion network (STCnet) to explicitly tackle the partial occlusion problem by recovering the appearance of the occluded parts. For one thing, according to the spatial structure of pedestrian frame, the visible (unoccluded) body parts can be used to predict the missing (occluded) body part of a person. For another, because of the temporal patterns of pedestrian sequence, the information from adjacent frames is helpful for recovering the appearance of the current frame. Motivated by the two facts, we design the spatial structure generator and temporal attention generator in STCnet. The spatial structure generator exploits the spatial information of the frame to predict the appearance of the occluded parts. The temporal attention generator exploits the temporal information of the video with a novel temporal attention layer to refine the parts generated by the spatial generator. With the spatial and temporal generators, STCnet is able to recover the occluded parts.  

Furthermore, we propose an occlusion-free video re-ID framework by combining a re-ID network with STCnet (VRSTC), where the unoccluded frames are used to train and test the re-ID network.  Due to the superior completion ability of STCnet, the video re-ID framework, VRSTC, achieves robustness to partial occlusion. We demonstrate the effectiveness of the proposed framework on three challenging video re-ID datasets, and our method outperforms the state-of-art methods under multiple evaluation metrics.

\section{Related Works}
\textbf{Person re-identification.} 
Person re-ID for still images has been extensively studied \cite{img1,IVC,LADF,img3,img4,img5,PCB}. Recently, researchers start to pay attention to video re-ID \cite{STFV3D,RCN,re-conv,temporal-enhanced,See,diversity,jointly,QAN,RQAN}. McLaughlin \etal \cite{RCN} and Wu \etal \cite{re-conv} proposed a basic pipeline for deep video re-ID. First, the frame features are extracted by convolutional neural network. Then a recurrent layer is applied to incorporate temporal context information into each frame. Finally, the temporal average pooling is adopted to obtain video representation. Wu \etal \cite{temporal-enhanced} further proposed a temporal convolutional subnet to extract local motion information. These methods verify that the temporal information of video can help to identify the person. However, because these methods treat each frame of video equally, the frames with partial occlusion will distort the video representation.

To handle partial occlusion, the attention based approaches are gaining popularity. Zhou \etal \cite{See} proposed a RNN temporal attention mechanism to select the most discriminative frames from video. Liu \etal \cite{QAN} used a convolutional subnet to predict quality score for each frame of video. Xu \etal \cite{jointly} presented a Spatial and Temporal Attention Pooling Network, where the spatial attention pooling layer selected discriminative regions from each frame and the temporal attention pooling selected informative frames in the sequence. Similarly, Li \etal \cite{diversity} used multiple spatial attention modules to localize distinctive body parts of person, and pooled these extracted local features across time with temporal attention.

Overall, the above methods process partial occlusion problem by discarding the occluded parts, which results in the loss of spatial and temporal information of video. Different from the existing methods, we explicitly tackle the partial occlusion problem by recovering the occluded parts. Then the recovered parts are leveraged together with the unoccluded parts for robust video reID under partial occlusion.

\begin{figure*}[t]
\begin{center}
\captionsetup{font={small}}
\centerline{\includegraphics[width=0.8\textwidth]{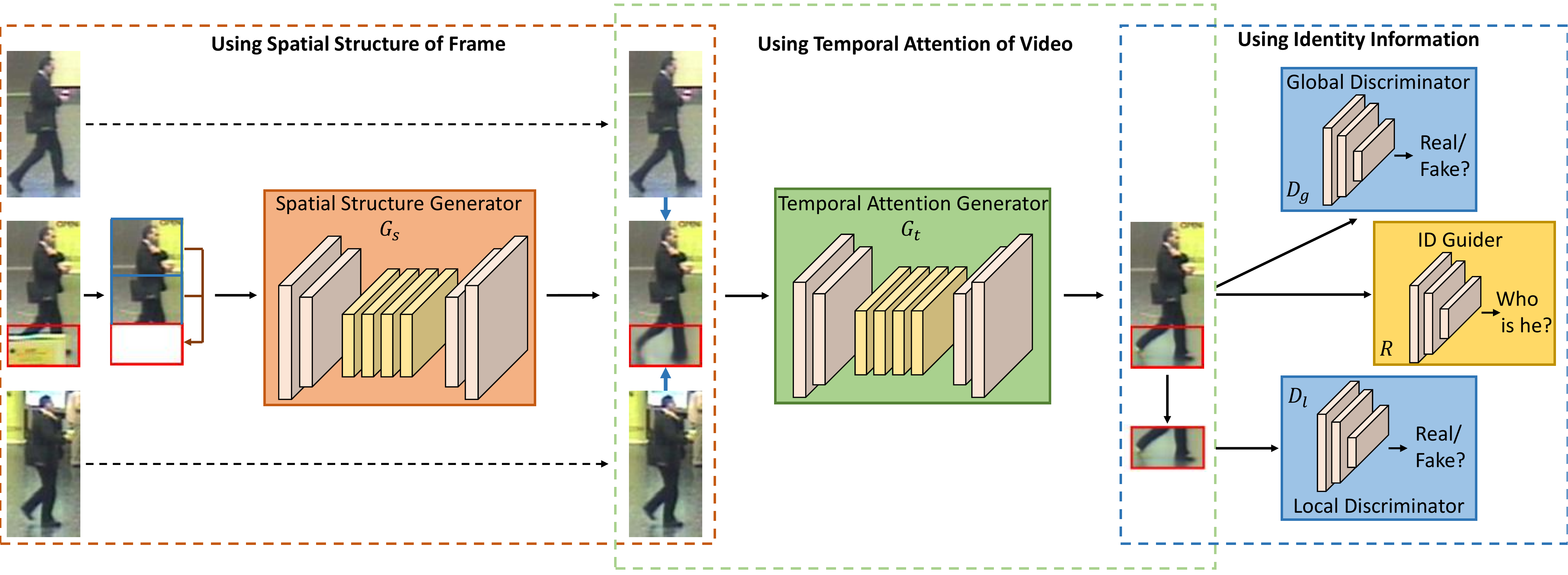}}
\caption{Overview of STCnet. The spatial structure generator takes the masked frame as input and outputs the generated frame. The temporal attention generator refines the generated frame with the adjacent frames. Two discriminators distinguish the synthesize contents in the mask and whole generated frame as real and fake. The ID guider network is to ensure the identity of the generated frame.}
\label{architecture}
\end{center}
\vspace{-8mm}
\end{figure*}

\textbf{Image completion.}
Image completion aims to fill the missing or masked regions in images with plausibly synthesized contents. It has many applications in photo editing, textual synthesis and computational photography. Early works \cite{inpaint1,inpaint2} attempted to solve the problem by matching and copying background patches into the missing regions. Recently, deep learning approaches based on Generative Adversarial Network (GAN) \cite{GAN} had emerged as a promising paradigm for image completion.  Pathak \etal \cite{context} proposed Context Encoder that generated the contents of an arbitrary image region conditioned on its surroundings. It was trained with pixel-wise reconstruction and an adversarial loss, which produced sharper results than training the model with only reconstruction loss.  Iizuka \etal \cite{globally} improved  \cite{context} by using dilated convolution \cite{dilated} to handle arbitrary resolutions. In \cite{globally}, global and local discriminators were introduced as adversarial losses. The global discriminator pursued global consistency of the input image, while the local discriminator encouraged the generated parts to be valid. Our proposed STCnet builds on \cite{globally} and extends it to exploit the temporal information of video by the proposed temporal attention module. In addition, STCnet employs a guider sub-network endowed with a re-ID cross-entropy loss to preserve the identities of the generated images.

\section{Spatial-Temporal Completion network}
In this section, we will first illustrate the overview of the proposed STCnet. Then we will demonstrate the details about each module of STCnet. Finally, the objective function to optimize STCnet will be given. 

\subsection{Network Overview} 
The key idea of STCnet is to alleviate the interference of occluders on the extracted features for pedestrian retrieval via explicitly recovering the occluded parts with spatio-temporal information of video. The network architecture of STCnet is shown in Figure \ref{architecture}.

STCnet consists of spatial structure generator, temporal attention generator, two discriminators and an ID guider subnetwork. The spatial structure generator leverages the spatial structure of the pedestrian frame, and makes an initial coarse prediction for the contents of occluded parts conditioned on the visible parts of this frame. The temporal attention generator takes use of the temporal patterns of the video, and refines the contents of the occluded parts with the information from adjacent frames. We introduce a local discriminator for the occluded regions to generate more realistic results, and a global discriminator for the entire frame to pursue the global consistency. In addition, an ID guider subnetwork is adopted to preserve the ID label of the frame after completion.

\subsection{Spatial Structure Generator}
Because of spatial structure of the frames in pedestrian video, the contents of occluded parts can be predicted with the visible parts of the frames. To the end, we design the spatial structure generator to model the correlation between the occluded and visible parts. 

Spatial structure generator is designed as an autoencoder. The encoder takes a frame with white pixels filled in the occluded parts (all the pixels in the occluded regions are set to $0$) as input, which is denoted as masked frame, and produces a latent feature representation of this frame. The decoder takes the feature representation and generates the contents for the occluded parts. In addition, we adopt the dilated convolution \cite{dilated} in the encoder to enlarge the size of the receptive fields, which can help to propagate the information from distant visible parts to the occluded parts. 

The architecture of spatial structure generator is derived from the completion network \cite{globally}. In term of layer implementations, we use the convolution with $3\times 3$ kernels and ELUs \cite{elu} as activation functions. The encoder consists of five convolutional layers and stacks four dilated convolutional layers of that, which decreases the resolution to a quarter of the original size of the input frame. The decoder consists of two deconvolution layers \cite{deconvolution} to restore the original resolution of the frame. 

\subsection{Temporal Attention Generator}
\begin{figure}[t]
\captionsetup{font={small}}
\centerline{\includegraphics[width=1.0\columnwidth]{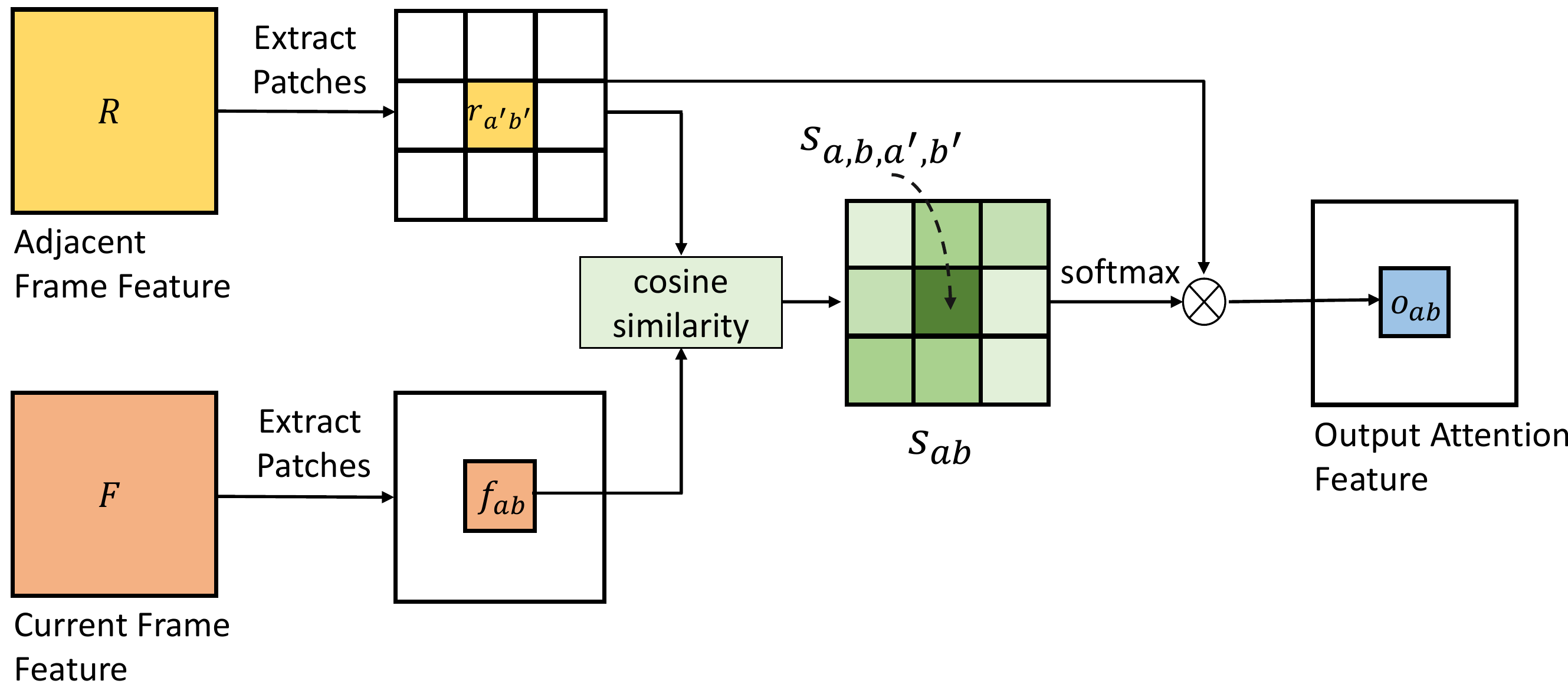}}
\caption{Illustration of the temporal attention layer. For simplicity, we only describe the generation process of one patch  ($o_{a,b}$) of output feature. The generation process of other patches is similar.}
\label{attention}
\vspace{-3mm}
\end{figure}

In view of the temporal patterns of video, the information from adjacent frames can also be exploited to predict the contents of the occluded parts. So we introduce a novel temporal attention layer, which learns where to attend feature from adjacent frames to generate the contents of the occluded parts. It is differentiable and can be integrated into the temporal attention generator.

The temporal attention layer is able to model relationships between the generated frames of spatial generator and the adjacent frames. For simplicity, we denote the generated frames of spatial generator as current frames. As shown in Figure \ref{attention}, we first extract patches ($3\times 3$) in the current frame feature ($F$) and adjacent frame feature ($R$). Then, we measure the normalized inner product (cosine similarity) between the patch of $F$ and the patch of $R$:
\begin{equation}
s_{a,b,a',b'} = \langle \frac{f_{a,b}}{||f_{a,b}||_{2}}, \frac{r_{a',b'}}{||r_{a',b'}||_{2}} \rangle,
\end{equation} 
where $f_{a,b}$ denotes the patch centered at location $\left(a,b\right)$ in current frame, $r_{a',b'}$ denotes the patch centered at location $\left(a',b'\right)$ in adjacent frame, $s_{a,b,a',b'}$ indicates similarity between $f_{a,b}$ and $r_{a',b'}$. Then we normalize the similarity with the softmax function:
\begin{equation}
s^{*}_{a,b,a',b'}=\frac{\exp (s_{a,b,a',b'})}{\sum_{c'd'}\exp (s_{a,b,c',d'})}.
\end{equation}
Finally, for each patch of current frame, it is updated via aggregating all patches of adjacent frames with weighted summation, where the weighs are decided by the similarity between the corresponding two patches:
\begin{equation}
o_{a,b} = \sum_{a'b'} s^{*}_{a,b,a',b'}r(a',b').
\end{equation}

To integrate temporal attention layer, we introduce three parallel encoders in the temporal attention generator. An encoder for the occluded frame focuses on hallucinating contents, while the other two encoders are for precious and next adjacent unoccluded frame receptively. Two temporal attention layers are appended on top of the encoders to attend on adjacent frames features of interest. Output features from three encoders are then concatenated and fed into a decoder to obtain the final output. The architectures of the encoders and decoder of the temporal generator are the same as those in the spatial generator.

\subsection{Discriminator}
We adopt a local and a global discriminator to improve the quality of generated contents of the occluded parts. The local discriminator takes the occluded parts as inputs and determines whether the synthesized contents in the occluded parts are real or not. It helps to generate detailed appearance and encourages the generated parts to be valid. The global discriminator takes the entire frames as inputs and regularizes the global structure of the frames. The two discriminators work collaboratively to ensure that the generated contents of occluded parts are not only realistic, but also consistent with surrounding contexts.   

The architecture of the two discriminators is similar to \cite{dcgan}, which consists of six convolutional layers and a single fully-connected layer. All the convolutional layers use $3\times3$ kernels and a stride of $2\times2$ pixels to decrease the frame resolution. The fully-connected layer uses sigmoid as activation function, which outputs the probability that the input is real. 

\subsection{ID Guider}
\label{ID_Guider}
In order to make the completed (unoccluded) frames boost the person re-ID performance, we introduce an ID guider subnetwork to guide the generators more adapted to re-ID problem. The ID guider subnetwork takes in the completed frames and output the classification results which are forced to be the real categories. In this way, the identity cues of the person are preserved during completion.

We employ ResNet-50 \cite{resnet-50} as the backbone network and modify the output dimension of the classification layer to the number of training identities. Following \cite{PCB}, we remove the last spatial down-sampling operation in the ResNet-50 to increase retrieval accuracy with very light computation cost added.

\subsection{Object Function}
STCnet is trained with three loss functions jointly: a reconstruction loss to capture the overall structure, an adversarial loss to improve the realness, and a guider loss to preserve the ID of the generated frames. Notably, we replace pixels in the non-mask (unoccluded) region of generated frames with original pixels.

We first introduce the reconstruction loss $L_{r}$ for the spatial generator $G_{s}$ and temporal generator $G_{t}$, which is the $L_{1}$ distances between the network output and the original frame:
\begin{equation}
L_{r} = ||x - \hat{x}_1||_1 + ||x - \hat{x}_2||_1 
\end{equation}
\begin{equation}
\hat{x}_1 = M \odot G_{s}((1-M)\odot x) + (1 - M) \odot x
\end{equation}
\begin{equation}
\hat{x}_2 = M \odot G_{t}(\hat{x}_1, x_{p}, x_{n}) + (1 - M) \odot x
\end{equation}
where $x$ is the input of the spatial generator, $x_{p}$ and $x_{n}$ are previous and next adjacent frames of $x$ respectively, $\hat{x}_1$ and $\hat{x}_2$ are the predictions of the spatial and temporal generators respectively, $M$ is a binary mask corresponding to the dropped frame region with value $1$ wherever a pixel was dropped and $0$ for elsewhere, and $\odot$ is the element-wise product operation.

With the global discriminator $D_{g}$ and local discriminator $D_{l}$, we define a global adversarial loss $L_{a_1}$ which reflects the faithfulness of the entire frame, and a local adversarial loss $L_{a_2}$ which reflects the validity of the generated contents in the occluded part:
\begin{equation}
\begin{split}
L_{a_1} = \mathop{\min}\limits_{G_s, G_t}\mathop{\max}\limits_{D_g}\mathbb{E}_{x\sim p_{data}(x)}[\log D_{g}(x) \\
 + \log D_{g}(1-\hat{x}_2)]
\end{split}
\end{equation}
\begin{equation}
\begin{split}
L_{a_2} = \mathop{\min}\limits_{G_s, G_t}\mathop{\max}\limits_{D_l}\mathbb{E}_{x\sim p_{data}(x)}[\log D_{l}(M \odot x) \\
 + \log D_{l}(1- M \odot \hat{x}_2)]
\end{split}
\end{equation} 
where $P_{data}(x)$ represents the distribution of real frame $x$. 

As for the ID guider network $R$, the guider loss $L_c$ is the simple cross-entropy loss, which is expressed as:
\begin{equation}
L_c = -\sum_{k=1}^{K} q_k\log R(\hat{x}_2)_k
\end{equation} 
where $K$ is the number of classes and $q$ is the ground truth distribution of the input frames.

Finally, the overall loss function is defined by:
\begin{equation}
L = L_r + \lambda_{1}(L_{a_1} + L_{a_2}) + \lambda_{2}L_{c}
\end{equation} 
where $\lambda_{1}$ and $\lambda_{2}$ are the weights to balance the effects of different losses.

\section{Occlusion-Free Video Person Re-ID}

\begin{figure}[t]
\begin{center}
\centerline{\includegraphics[width=1.0\columnwidth]{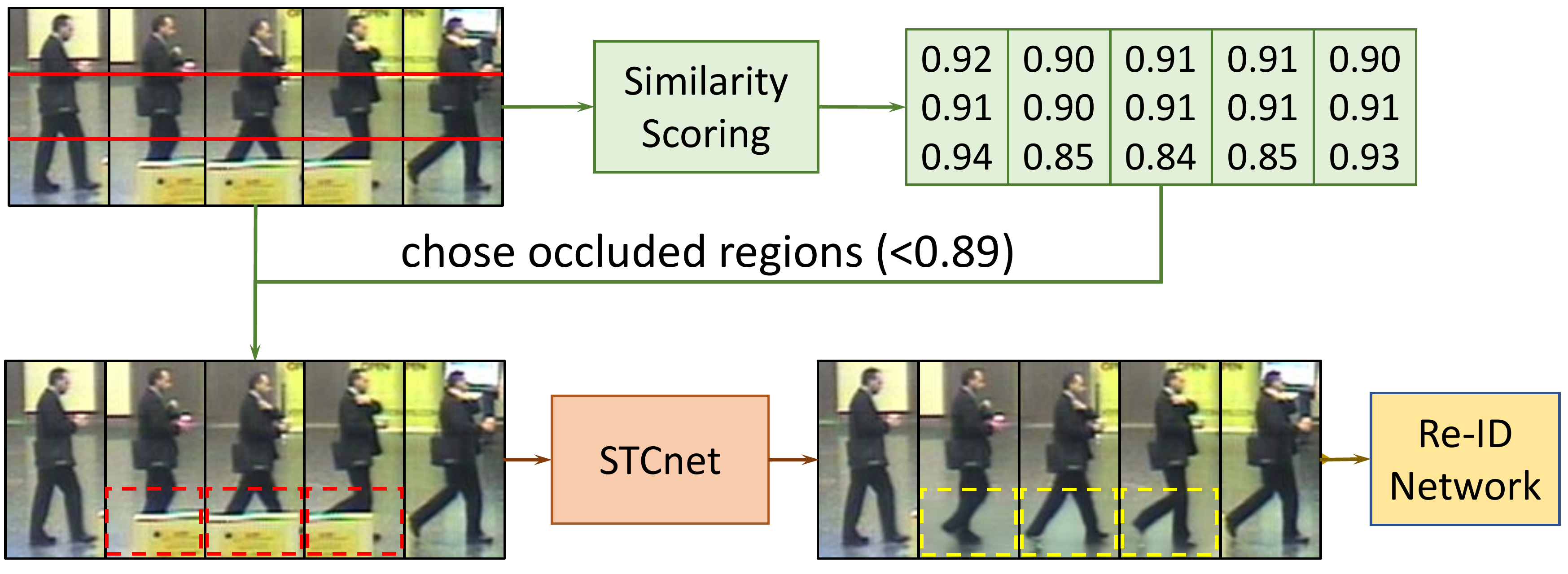}}
\captionsetup{font={small}}
\caption{Pipeline of VRSTC.}
\label{pipeline}
\end{center}
\vspace{-10mm}
\end{figure}

By combining STCnet with a re-ID network, we propose a video re-ID framework VRSTC, which is robust to partial occlusion. The framework of VRSTC is shown in Figure \ref{pipeline}. First, a similarity scoring mechanism is proposed to locate the occluded parts of frames. Then, STCnet is adopted to recover the appearance of the occluded parts. Finally, the recovered regions are leveraged with those unoccluded regions to train the re-ID network. Without designing complicated model and loss function, our framework can achieve great performance improvement. 

\subsection{Similarity Scoring}
The works \cite{QAN,See,jointly,snippet} use the attention mechanism to locate the occluded frames. These approaches usually construct a subnetwork to predict the weight of each frame in video. However, it is difficult for the subnetwork to automatically assign low weights to the occluded frames, as there is no direct supervision for the weights. 

Considering the concern above, we propose a similarity scoring mechanism to generate the attention score for each region of frames. Motivated by the observation that the occlusion usually occurs in a few consecutive frames and the occluders have different semantic features from the original body parts, we use the cosine similarity between the frame region feature and the video region feature as the score. Formally, we denote the input video as $I=\{I_{t}\}_{t=1}^{T}$, where $T$ indicates the length of the video. The frames are vertically divided into three fixed regions equally $I_{t}=\{I_t^u, I_t^m, I_t^l\}$, where $u$, $m$, and $l$ represent the upper, middle and lower part of the frames respectively. The feature representation of each region $\{v_t^k|k\in\{u,m,l\}\}$ is extracted using convolutional neural network. The video region feature is then obtained by average pooling according to temporal domain:

\begin{equation}
\overline{v}^{k}=\frac{1}{T}\sum_{t=1}^{T}v_{t}^{k}, \quad \text{where}\ k\in\{u,m,l\}
\end{equation} 

Next, the score of each frame region is calculated with the following equation: 

\begin{equation}
u_{t}^{k} = \left< \frac{v_t^k}{||v_t^k||_{2}}, \frac{\overline{v}^{k}}{||\overline{v}^{k}||_{2}} \right>
\end{equation}
 
 In the last, we regard those regions with scores lower than a threshold $\tau$ ($0.89$ in our work) as the occluded regions. We replace the occluded regions with the generated regions by STCnet to form a new dataset and train a re-ID network with the new dataset.

\subsection{Re-ID Network}
\label{reID}
Most re-ID networks and loss functions can combine with STCnet. Note that STCnet can combine with the most advanced re-ID models to further enhance the overall performance. In order to verify the effectiveness of STCnet as a kind of data enhancement method, we use a simple re-ID network with average temporal pooling and the cross-entropy loss. 

We employ the modified ResNet-50 as the backbone network. In order to capture temporal dependency, we embed the non-local blocks \cite{non-local} into the re-ID network. Different from the previous works that only build temporal dependency in the end, the non-local blocks can be inserted into the earlier part of deep neural networks. This allows us to build a richer hierarchical temporal dependency that combines both non-local and local information.

\section{Experiments}

\subsection{Datasets and evaluation protocols} 

\textbf{iLIDS-VID} dataset consists of $600$ video sequences, where $300$ different identities are captured by two cameras. Each video sequence contains $23$ to $192$ frames. 

\textbf{MARS} dataset is the largest video re-ID benchmark with $1,261$ identities and around $20,000$ video sequences captured from $6$ cameras. 
The bounding boxes are produced by DPM detector \cite{DPM} and GMMCP tracker \cite{Gmmcp}. 

\textbf{DukeMTMC-VideoReID} dataset is a subset of the tracking dataset DuKeMTMC \cite{duke} for video person re-ID. The pedestrian images are cropped from the videos for 12 frames every second to generate a tracklet. 

\textbf{Evaluation protocol:} We adopt mean Average Precision (mAP)~\cite{map} and Cumulative Matching Characteristics (CMC)~\cite{cmc} as evaluation metrics. 


\subsection{Implementation Details}
In this subsection, we give the implementation details of our approach. We use PyTorch \cite{pytorch} for all experiments.

\textbf{Pre-training a re-ID network.}  We train ResNet-50 with cross-entropy loss to be the ID guider of STCNet. In training term, four-frame input tracks are cropped out from an input sequence. The frame features are extracted by ResNet-50, then the average temporal pooling is used to obtain the sequence feature. Input images are resized to  $256\times 128$. The batch size is set to 32. For the data augmentation, we only use random horizontal mirroring for training. We adopt the Adaptive Moment Estimation (Adam) \cite{adam} with weight decay of $0.0005$. The network is trained for 150 epochs in total, with an initial learning rate of $0.0003$ and reduced it with decay rate $0.1$ every 50 epochs.

\textbf{Locating occluded regions.} With the pretrained re-ID network as feature extractor, we use the similarity scoring mechanism to generate the score for each frame region. We regard the regions whose scores are lower than $\tau$ as the occluded regions, and we define the frames without occluded regions as the unoccluded frames. In our experiment, $\tau$ is set to $0.89$.

\textbf{Training STCnet.} To train STCnet, we need to build a training set consisting of the input occluded frames and target de-occluded frames. However, there is no ground-truths for the occluded frames. So we only use the unoccluded frames from the training set of target re-ID dataset to train STCnet, and randomly mask a region of the unoccluded frames as inputs. The input and target frames are resized to $128\times 64$ and linearly scaled to $[-1,1]$. The parameters of ID guider are fixed when training STCnet. We optimize the spatial and temporal generators and two discriminators with alternating Adam optimizer, and the learning rate is set to $0.0001$. $\lambda_1$ and $\lambda_2$ are set to $0.001$ and $0.1$ respectively. Once the training is over, STCnet can recover the appearance of the occluded regions. 

\textbf{Improving re-ID network with de-occluded frames.} The occluded regions of the frames in raw re-ID dataset are replaced with the regions generated by STCnet to form a new dataset. Then the re-ID network is trained and tested with the new dataset. We embed the non-local block \cite{non-local} in the re-ID network to capture temporal dependency of input sequence. According to the experiments in \cite{non-local}, five non-local blocks are inserted to before the last residual block of a stage. Three blocks are inserted into $\textit{res}_4$ and two blocks are inserted into $\textit{res}_3$, to every other residual block. Other settings are the same as those in the experiments of pre-training a re-ID network. During testing, given an input of entire video, the video feature is extracted using the trained re-ID network for retrieval under Euclidean distance. 

\subsection{Ablation Study}
\label{ablation_study}

\subsubsection{Component Analysis of STCnet} 
We investigate the effect of each component of STCnet by conducting several analytic experiments. Table \ref{inpaint} reports the results of each component of STCnet. Baseline corresponds to ResNet-50 trained on raw target dataset. NL embeds the non-local blocks into the baseline model and improves the results, which indicates that non-local blocks are effective for integrating temporal information of video. In the other experiments of this part, we replace the occluded regions with generated regions by different completion models to form a new dataset and train and test NL on the new dataset.

\begin{table}[t]
\centering
\small
\captionsetup{font={small}}
\caption{Comparative analysis of STCnet. The rank-1 CMC accuracy is reported and mAP is reported for MARS and DukeMTMC-VideoReID in brackets.}
\vspace{-3mm}
\label{inpaint}
\begin{center}
\begin{tabular}{|l |c | c |c| }
\hline                   
Methods & iLIDS & MARS & DukeMTMC\\
\hline                      
baseline        &79.8 & 84.4 (77.2) & 91.4 (90.0)  \\
NL              &80.1 & 86.1 (79.9) & 91.8 (91.2)  \\
\hline
VRSTC      &\textbf{83.4}  &\textbf{88.5 (82.3)} &\textbf{95.0 (93.5)} \\
\hline
\end{tabular}
\end{center}
\vspace{-5mm}
\end{table}

\textbf{Spatial structure generator.} Spa denotes the spatial structure generator trained only with the spatial reconstruction loss. Compared with NL, Spa improves the rank-1 accuracy by $1.3\%$, $0.9\%$ and $1.1\%$ on iLIDS-VID, MARS and DukeMTMC-VideoReID respectively. The result shows that the spatial structure generator, which utilizes the spatial information of frames to recover the appearance of occluded regions, is useful for boosting re-ID performance. 
 
\textbf{Temporal attention generator.} Spa+Tem consists of spatial and temporal generators, which is trained with both spatial and temporal reconstruction loss. By comparing Spa and Spa+Tem, we can see that the proposed temporal generator further improves accuracy. We argue that the temporal attention layer can attend the information from adjacent frames, which makes the generated frames more semantically consistent with the video sequence. The re-ID network (NL) can then extract better temporal information of the resulting sequence, leading to a more discriminative video feature representation.  

It is noteworthy that the improvement of temporal generator does not come from the increased depth by naively adding extra layers to the spatial generator. To see this, we also try two variants of temporal generator: \textbf{Autoencoder (AE)} and \textbf{Temporal Autoencoder (TAE)}. AE is a standard autoencoder and only takes the predictions of spatial generator as inputs. It has the same encoder and decoder with temporal generator expect the number of filters in the encoder is tripled. This controls for the total number of parameters in AE compared to temporal generator. TAE is the temporal generator without the temporal attention layer. As shown in Table~\ref{inpaint}, Spa+AE does not increase the accuracy compared to Spa. This shows that the improvement of temporal generator is not because it adds extra layers to the spatial generator. In addition, the temporal generator performs better than TAE. This improvement shows that the proposed temporal attention layer makes better use of the temporal information to generate more discriminative frames.

\textbf{Discriminators.} Spa+Tem+LD consists of the two generators and the local discriminator. Spa+Tem+LD+GD further incorporates the global discriminator. Both are trained with reconstruction and adversarial losses. From the results, we can see that the discriminators only slightly improve the performance. We argue that the discriminators aim to generate more visually realistic frames, without bringing too much additional discriminant information for re-ID.  

\textbf{ID guider network.} The final model STCnet is trained with the reconstruction, adversarial and guider losses. The generated samples achieve better performance with the ID guider, which suggests that the ID guider is beneficial to generate suitable samples for training re-ID network. The improvement can be attributed to the ability of preserving the underlying visual cues associated with the ID labels.

\begin{figure}[t]
\begin{center}
\centerline{\includegraphics[width=1.0\columnwidth]{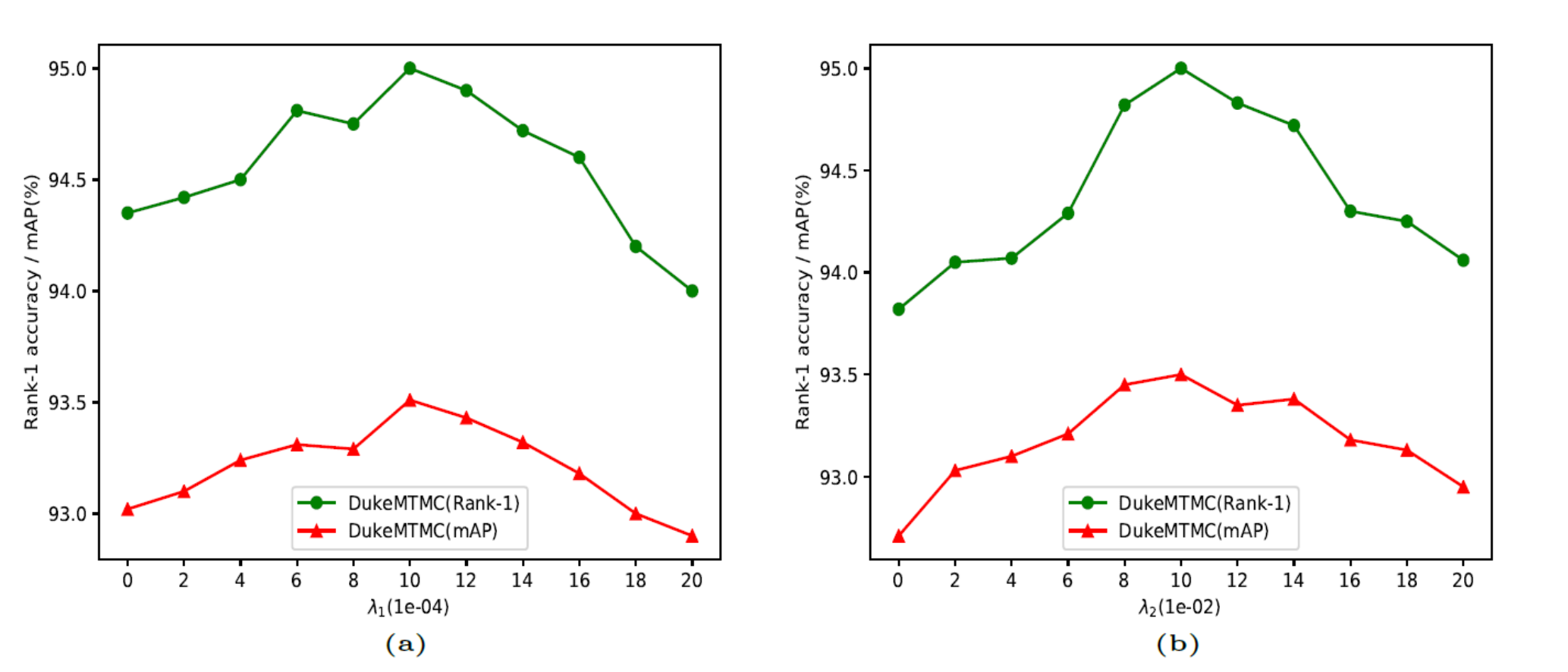}}
\captionsetup{font={small}}
\caption{The rank-1 and mAP on DukeMTMC-VideoReID (a) different $\lambda_{1}$ and fixed $\lambda_{2}$=0.1, (b) different $\lambda_{2}$ and fixed $\lambda_{1}$=0.001.}
\label{param}
\end{center}
\vspace{-8mm}
\end{figure}

\begin{table}[!t]
\centering
\small
\captionsetup{font={small}}
\caption{Comparison of different threshold $\tau$ in the similarity scoring mechanism. The rank-1 CMC accuracies are reported and mAP are reported for MARS and DukeMTMC-VideoReID in brackets.}
\label{threshold}
\vspace{-3mm}
\begin{center}
\begin{tabular}{l |c |c |c }
\hline
threshold ($\tau$) & iLIDS & MARS & DukeMTMC\\
\hline
0 (baseline)    &79.8 &84.4 (77.2) &91.4 (90.0) \\
0.88            &80.0 &84.8 (77.2) &91.5 (90.3) \\
0.89            &\textbf{80.3} &\textbf{84.9 (77.4)} &\textbf{91.7} (90.5) \\
0.91            &78.8 &84.2 (77.0) &91.4 (\textbf{90.6}) \\
0.93            &78.3 &83.6 (76.6) &91.4 (90.5) \\
\hline
\end{tabular}
\end{center}
\vspace{-2mm}
\end{table} 

\subsubsection{Influence of the parameters $\lambda_{1}$ and $\lambda_{2}$}
$\lambda_{1}$ and $\lambda_{2}$ are two parameters to balance the relative effects of the adversarial loss and guider loss respectively. We analyze the impact of the $\lambda_{1}$ and $\lambda_{2}$ on DukeMTMC-VideoReID, and the results are shown in Figure \ref{param} (a) and (b) respectively. We observe that our method achieves the best performance when $\lambda_{1}$ is set to 0.001 and $\lambda_{2}$ is set to 0.1. Notice that there will be a big performance degradation when $\lambda_{1}$ or $\lambda_{2}$ are too big. The main reason is that STCnet becomes difficult to converge if the adversarial loss or guider loss takes a dominant role.

\begin{figure}[!t]
\begin{center}
\centerline{\includegraphics[width=1.0\columnwidth]{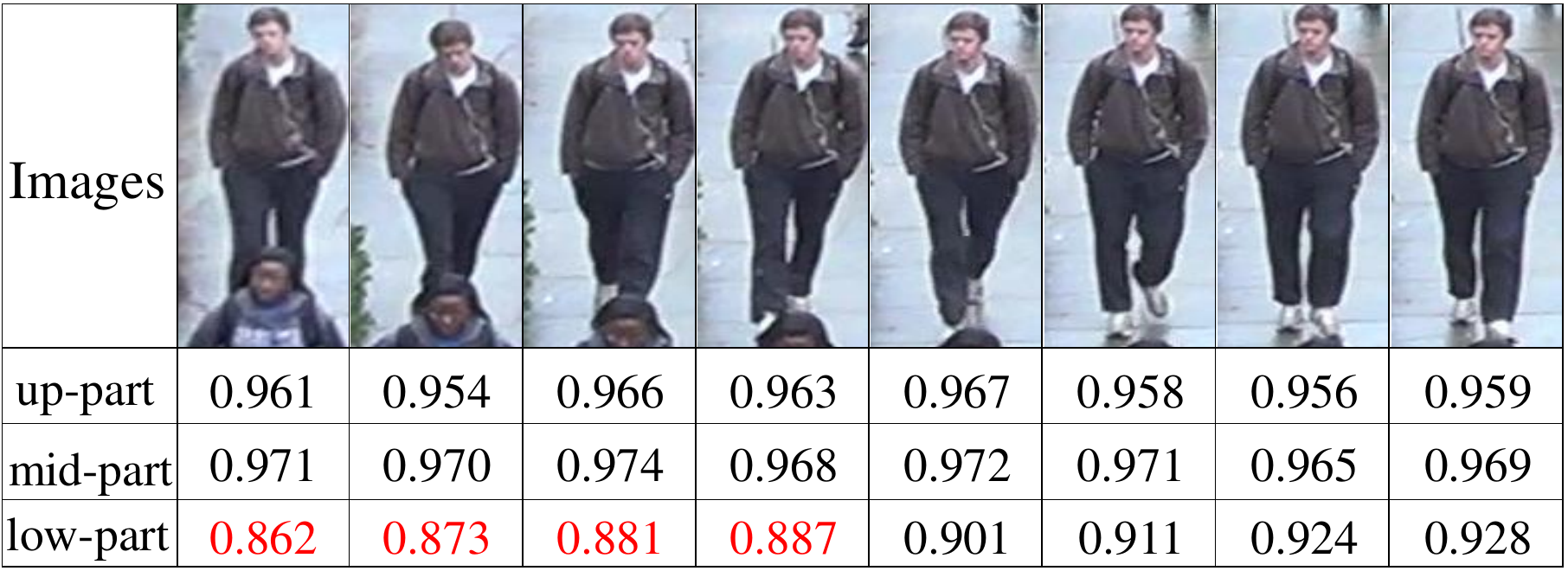}}
\captionsetup{font={small}}
\caption{Scores of similarity scoring mechanism from one sequence. Red represents small score.}
\label{score}
\end{center}
\vspace{-10mm}
\end{figure}

\subsubsection{Influence of the threshold $\tau$} 
We also carry out experiments to investigate the effect of varying the threshold $\tau$ in the similarity scoring mechanism. The experiment setting is as follows. Giving an input video sequence, we first discard the frames with occluded regions whose scores are lower than $\tau$. The video feature is then obtained with the remaining frames using average temporal pooling. Finally, we use the obtained video feature to compute the similarity between videos under Euclidean distance. Notably, when $\tau=0$, we keep all frames of a video, which is the same as the baseline model. 

As shown in Table \ref{threshold}, there is an improvement in performance when $\tau$ is increased, which implies that the discarded frames would have corrupted the representation of video. This result implicitly demonstrates the scores achieved by similarity scoring can locate occluded frames. However, as $\tau$ is further increased, the accuracy drops gradually. The main reason is that the unoccluded frames may be discarded with too large threshold. The network achieves the best performance when $\tau=0.89$. So we set $\tau$ to $0.89$ in our experiments. Notably, the introduction of the frames completed by STCnet can further improve the performance (see Table \ref{inpaint}), which demonstrates that the contents restored by STCnet can help identify the person.

In order to demonstrate the similarity scoring mechanism more intuitively, the scores of one sequence from DukeMTMC-VideoReID is visualized in Figure \ref{score}. 
Due to the occlusion from another person, the scores of the lower parts of the first four frames are relatively small. This results further demonstrate the score achieved by the proposed similarity scoring mechanism can reflect the visibility of each region. 

\subsection{Comparison with State-of-the-arts} 
\label{state-of-the-arts}

\begin{table}[t]
\small
\centering
\captionsetup{font={small}}
\caption{Comparison with related methods on MARS. * denotes those requiring optical flow as inputs.}
\label{mars}
\begin{center}
\begin{tabular}{|l | c c c c|}
\hline
\multirow{2}*{Methods} &\multicolumn{4}{|c|}{MARS} \\
\cline{2-5}
&rank-1 &rank-5 &rank-10 &mAP \\
\hline
QAN          \cite{QAN}    &73.7 &84.9 &91.6 &51.7 \\
K-reciprocal  \cite{k-reciprocal}   &73.9 &-    &-    &68.5 \\
RQEN        \cite{RQAN}     &77.8 &88.8 &94.3 &71.7 \\
TriNet       \cite{TriNet}    &79.8 &91.4 &-    &67.7 \\
EUG \cite{dukereid} &80.8 &92.1 &96.1 & 67.4 \\
STAN         \cite{diversity}    &82.3 &-    &-    &65.8 \\
Snipped \cite{snippet}            &81.2 &92.1 &-    &69.4 \\
Snippet+OF*       \cite{snippet}   &86.3 &94.7 &\textbf{98.2} &76.1 \\
\hline
VRSTC       &\textbf{88.5} &\textbf{96.5}  &97.4 &\textbf{82.3}\\
\hline
\end{tabular}
\end{center}
\vspace{-3mm}
\end{table}

\begin{table}[!t]
\small
\centering
\captionsetup{font={small}}
\caption{Comparison with methods on DukeMTMC-VideoReID.}
\label{duke}
\vspace{-3mm}
\begin{center}
\begin{tabular}{l | c c c c}
\hline
\multirow{2}*{Methods} &\multicolumn{4}{c}{DukeMTMC} \\
\cline{2-5}
&rank-1 &rank-5 &rank-10 &mAP \\
\hline
EUG \cite{dukereid} &83.6 &94.6 &97.6 &78.3 \\
\hline
VRSTC &\textbf{95.0} &\textbf{99.1} &\textbf{99.4} &\textbf{93.5} \\
\hline
\end{tabular}
\end{center}
\vspace{-2mm}
\end{table}

\begin{table}[!t]
\small
\centering
\captionsetup{font={small}}
\caption{Comparison with related methods on iLIDS-VID.}
\label{ilids}
\vspace{-2mm}
\begin{center}
\begin{tabular}{l | c c c c}
\hline
\multirow{2}*{Methods} &\multicolumn{4}{c}{iLIDS-VID} \\
\cline{2-5}
&rank-1 &rank-5 &rank-10 &rank-20 \\
\hline
LFDA \cite{LFDA}      &32.9 &68.5 &82.2 &92.6 \\
KISSME \cite{KISSME}    &36.5 &67.8 &78.8 &87.1 \\
LADF \cite{LADF}      &39.0 &76.8 &89.0 &96.8 \\
STFV3D  \cite{STFV3D}   &44.3 &71.7 &83.7 &91.7 \\
TDL   \cite{TDL}     &56.3 &87.6 &95.6 &98.3 \\
\hline
Mars  \cite{mars} &53.0 &81.4 &-    &95.1 \\
SeeForest \cite{See}   &55.2 &86.5 &-    &97.0 \\
CNN+RNN* \cite{RCN}   &58.0 &84.0 &91.0 &96.0 \\
Seq-Decision \cite{sequence-decision} &60.2 &84.7 &91.7 &95.2 \\
ASTPN*   \cite{jointly}   &62.0 &86.0 &94.0 &98.0 \\
QAN   \cite{QAN}     &68.0 &86.8 &95.4 &97.4 \\
RQEN   \cite{RQAN}    &77.1 &93.2 &97.7 &99.4 \\
STAN \cite{diversity}    &80.2 &-    &-    &- \\
Snippet \cite{snippet}    &79.8   &91.8 &- &- \\
Snippet+OF* \cite{snippet}   &\textbf{85.4} &\textbf{96.7} &\textbf{98.8} &\textbf{99.5} \\
\hline
VRSTC     &83.4 &95.5 &97.7 &\textbf{99.5}  \\
\hline
\end{tabular}
\end{center}
\vspace{-3mm}
\end{table}

Table \ref{mars}, \ref{duke} and \ref{ilids} report the performance of our approach and other state-of-the-art methods on MARS, DukeMTMC-VideoReID and iLIDS-VID, respectively. On MARS and DukeMTMC-VideoReID, our approach outperforms the best existing methods. We attribute the improvements to the recovered contents of the occluded parts. The effective combination with STCnet makes our approach superior than the methods which only use the raw dataset. It is worth noting that DukeMTMC-VideoReID is recently proposed by \cite{dukereid} and our baseline model has outperformed \cite{dukereid} by $7.8\%$ and $11.7\%$ on rank-1 and mAP respectively. We hope it will serve as a new baseline on DukeMTMC-VideoReID.  On iLIDS-VID, our approach achieves slightly lower performance than Snippet+OF \cite{snippet}. Note that Snipper+OF uses additional optical flow as input to provide motion features, which is not utilized in our framework. In addition, our approach outperforms Snippet (without optical flow) significantly, which is a more fair comparison.

\subsection{Visualizing the effect of STCnet} 

\begin{figure}[t]
\begin{center}
\centerline{\includegraphics[width=0.8\columnwidth]{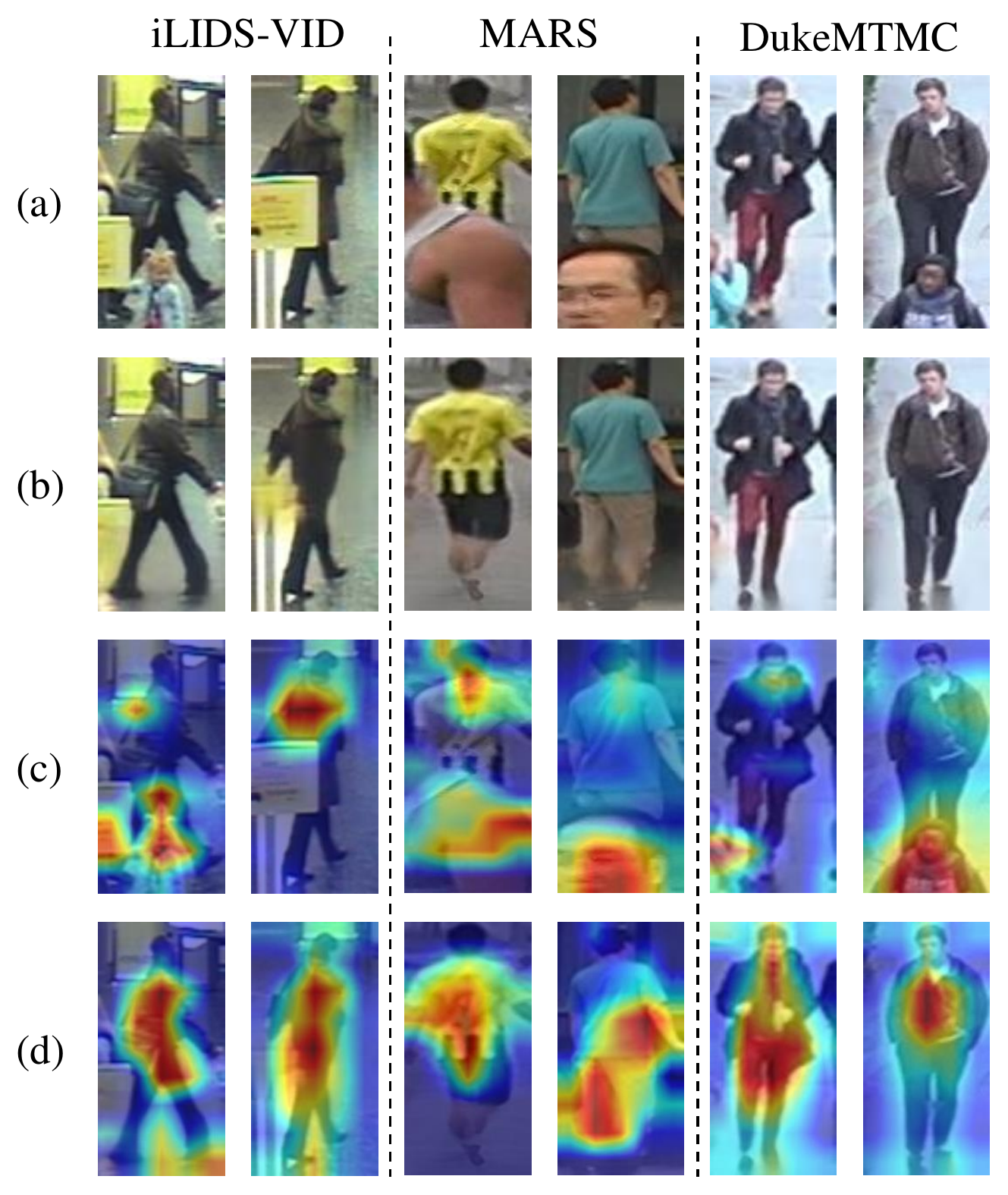}}
\captionsetup{font={small}}
\caption{Visual examples of STCnet. From top to bottom: (a) original image, (b) output of STCnet, (c) the activation maps of original image (d) the activation maps of completed image. Warmer color with higher value}
\label{visualize}
\end{center}
\vspace{-10mm}
\end{figure}

We visualize the generated frames of STCnet for intuitive exploration. Some partially occluded images are selected for evaluation. Figure \ref{visualize} provides a vivid illustration how STCnet recovers the contents of occluded parts and improves the extracted features. 
Specifically, when a person is occluded by some body part of other pedestrians, the feature representation extracted for the person is often corrupted by the visual appearances of the other pedestrians. As shown in the sixth column of Figure \ref{visualize} (c), the part of other pedestrians is activated by the re-ID network, which harms the feature representation of the target person. In addition, when a person is occluded by the environmental objects such as indicator and bicycles, there will be severe loss of body information in the feature extracted from the person (\eg the second column of Figure \ref{visualize} (c)). On the contrary, once STCnet recovers the contents of the occluded regions, the re-ID model will take more effective regions into account and discover new discriminative clues therefrom to recognize the person more correctly.

\section{Conclusion}
In this work, we present a novel framework combined a re-ID network with a completion network STCnet for video re-ID under partial occlusion. Aiming at explicitly tackling the partial occlusion problem, we design the STCnet to recover the appearance for the occluded regions and leverage the recovered regions with the unoccluded regions to train the re-ID network. Experiments on three datasets show that the proposed method outperforms the state-of-the-art video re-ID approaches. 

In the future, we will explore other types of deep generative architectures for recovering the appearance for the frames with extremely severe occlusion.

\noindent\textbf{Acknowledgement} 
This work is partially supported by National Key R\&D Program of China (No. 2017YFA0700800), Natural Science Foundation of
China (NSFC): 61876171, 61702486 and 6180618.

{\small
\bibliographystyle{ieee}
\bibliography{egbib}
}

\end{document}